%% file: main.tex
\documentclass[10pt,twocolumn,letterpaper]{article}
\usepackage{comment}
\usepackage[pagenumbers]{cvpr}

\usepackage{booktabs}
\usepackage{tabularx}
\usepackage{graphicx}
\usepackage{subcaption}
\usepackage{float}
\usepackage{rotating}
\usepackage{colortbl}
\usepackage{xcolor}
\usepackage{makecell}
\usepackage{caption}
\usepackage{comment}

\definecolor{cvprblue}{rgb}{0.21,0.49,0.74}
\usepackage[pagebackref,breaklinks,colorlinks,allcolors=cvprblue]{hyperref}

\def\paperID{12}
\def\confName{CVPR}
\def\confYear{2026}

\title{VISER: Visually-Informed System for Enhanced Robustness \\in Iris Presentation Attack Detection}

\author{Byron Dowling$^\dag$, Jacob Piland$^\dag$, Eleanor Frederick$^\dag$, Christopher Sweet$^\ddag$ and Adam Czajka$^\dag$\\
$^\dag$Department of Computer Science and Engineering, $^\ddag$Center for Research Computing\\
University of Notre Dame, Notre Dame, Indiana 46556, USA\\
{\small Corresponding emails: \tt\{bdowlin2,aczajka\}@nd.edu}
}

\begin{document}

\maketitle

\begin{figure*}[!ht]
    \centering
    \includegraphics[width=\linewidth]{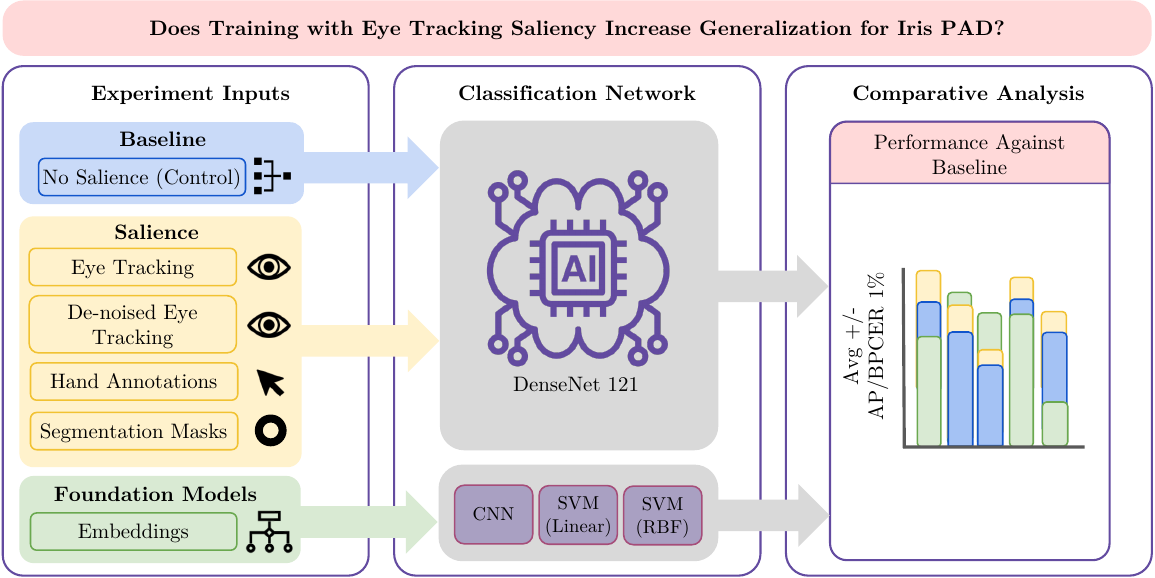}
    \caption{Experimental pipeline and paper contributions. We start with a DenseNet121 architecture with no saliency as our baseline. Different forms of saliency are then tested on the architecture using saliency-guided training. Next foundation model embeddings are tested in SVM-Linear, SVM-RBF, and CNN classifiers. Finally, AP/BPCER 1\% results are expressed as +/- improvement over the DenseNet121 baseline.}
    \label{fig:teaser}
\end{figure*}

\begin{abstract}
Human perceptual priors have shown promise in saliency-guided deep learning training, particularly in the domain of iris presentation attack detection (PAD). Common saliency approaches include hand annotations obtained via mouse clicks and eye gaze heatmaps derived from eye tracking data. However, the most effective form of human saliency for raising generalization to unknown attack classes in iris PAD remains under-explored. In this paper, we conduct a series of experiments comparing hand annotations, eye tracking heatmaps, segmentation masks, and foundation model embeddings to a state-of-the-art deep learning-based baseline on the task of unknown attack type classification for iris PAD. Results in a leave-one-attack-type out paradigm indicate that denoised eye tracking heatmaps show the best generalization improvement over cross entropy in Attack Presentation Classification Error Rate (APCER) at Bona Fide Presentation Classification Error Rate (BPCER) of 1\%. Along with this paper, we offer trained models, code, and saliency maps for reproducibility and to facilitate follow-up research efforts.
\end{abstract}

\section{Introduction}

Artificial Intelligence (AI) systems continue to play an increasingly pivotal role in everyday tasks, and researchers and companies alike understand that a lack of trust in an AI model or a lack of trust in a model's validity is a barrier to wide-scale adoption, particularly in critical tasks \cite{hasija2022artificial}. Among the core principles of Trustworthy AI is the idea of robustness, which means that good models should generalize well to unseen data classes \cite{li2023trustworthy, hasija2022artificial, kaur2022trustworthy}. One popular technique for increasing a model's ability to generalize to unknown attack types is utilizing human saliency (a map of features supporting the decision of a human examiner).

The use of human saliency in saliency-guided deep learning has seen increased interest in domains like synthetic face detection, x-ray abnormality detection, and iris PAD. Human saliency is used to improve the quality of model's feature maps \cite{dang2020detection}, and to directly guide the learning process of the model \cite{boyd2023cyborg}. These techniques have shown to increase a model's ability to generalize to unseen data types leading to more trustworthy and accurate performance. However, collecting human saliency can be expensive and time-consuming. While attempts to maximize the number of saliency samples needed for the effective training of models has been explored \cite{crum2023teaching, crum2025mentor}, there is still the question of what \textit{form} of human saliency is optimal for the best transfer of expert's knowledge into deep learning. Finally, with the rapid evolution of highly-capable foundation models, it raises the question of whether experts and human saliency are needed to push the generalization abilities of PAD models.

Thus in this paper we present VISER, a Visually Informed System for Enhanced Robustness in iris PAD. VISER attempts to address the trade-off between the continuous but noisy signal that eye tracking provides and the information loss that occurs in translation from visual to motor saliency. Results show that eye tracking heatmaps improve over the DenseNet121 cross entropy baseline for average APCER at BPCER of 1\% with denoised eye tracking showing the best performance of the variants. The {\bf main contributions of this paper} are as follows:
\begin{itemize}
    \item experiments evaluating iris PAD models trained with different types of human saliency in the task of unknown attack type classification,
    \item a comparison of human saliency-guided models vs. foundation models in iris PAD unknown attack type classification,
    \item a method of denoising raw eye tracking gaze maps for the purposes of saliency-guided training leading to a state-of-the-art iris PAD generalization performance.
\end{itemize}

This paper is organized around the following research questions:

\begin{itemize}
    \item[\textbf{RQ1:}] Does the use of eye tracking saliency in saliency-guided training generalize better to unseen data than other saliency types?
    \item[\textbf{RQ2:}] Does saliency-guided training outperform modern PAD solutions utilizing foundational models in iris PAD unknown attack type classification?
\end{itemize}

To facilitate follow-up research efforts, we offer the source codes, model weights, and saliency maps along with this paper for reproducibility purposes\footnote{\url{https://github.com/CVRL/VISER}}.

\section{Related Work}

\subsection{Detection of Biometric Presentation Attacks}

Iris recognition, widely regarded as a highly secure biometric modality, remains deployed in critical applications including border control and personal identification \cite{boyd2020iris, czajka2018presentation, daugman2009iris}. Recent developments highlight financial authentication systems as an emerging application for iris biometrics, illustrating its expanding role for secure payments \cite{PayEye2025}. As with any security-critical system, iris recognition remains vulnerable to adversarial attempts that seek to compromise its integrity \cite{czajka2018presentation}.

These attacks, known as presentation attacks, aim to manipulate a biometric system into an incorrect decision in one of two forms: identity concealment, where a malicious actor wishes to evade detection, or impersonation, where an actor attempts to obtain a false positive match \cite{boyd2023comprehensive, czajka2018presentation}. Iris presentation attacks include synthetic iris imagery generated using Generative Adversarial Networks (GANs) \cite{karras2020analyzing, karras2021alias, tinsley2023iris}, as well as physical presentation attacks involving direct sensor interaction, such as post-mortem cadaver samples, high-resolution iris printouts, and artificial eyes \cite{trokielewicz2018presentation, czajka2018presentation}. To combat these attacks, advances in deep learning have driven significant improvements in PAD systems \cite{Nguyen_CSUR_2024, boyd2023cyborg, crum2025mentor}. More recent work has explored leveraging the general vision capabilities of image foundation models \cite{Tapia_FG_2025, sony2025benchmarking} and designed novel loss functions to enhance multimodal alignment \cite{Zhang_ASIG_2024}.

\subsection{Human Saliency-Guided Training}

Machine learning techniques have been shown to exceed human performance in the context of biometrics tasks \cite{37o2012comparing}. However, alongside improving performance, there is growing demand for explainability which can be addressed by comparing model behavior to human experts \cite{42richardwebster2018visual}.
Previous work has seen human salience used in tasks beyond iris PAD: attention mechanisms \cite{31linsley2018learning}, natural language processing \cite{53zhang2020human}, scene description \cite{20he2019human,23huang2021specific}, handwriting, \cite{18grieggs2021measuring}, and in the biometric task of synthetic face detection where machine saliency has been shown to complement human expertise \cite{boyd2023cyborg,piland2023droid,35moreira2019performance,49trokielewicz2019perception}.
Despite this broad adoption, to the best of our knowledge, no prior work has directly compared human derived saliency (\eg, hand annotations and eye tracking data) and non-human saliency (\eg, segmentation maps) for iris PAD. 

\subsection{Visual vs Motor Saliency}
\label{sec:Visual vs Motor Saliency}

Human saliency can be collected in a variety of forms, with hand annotations captured via mouse clicks and eye tracking from commercial eye tracking devices being two of the most prominent examples. Despite their popularity, these approaches differ substantially from a cognitive perspective in terms of cognitive load and the processing of information and visual stimuli \cite{just1980theory, henderson2003human}. The distinction between the two saliency types is deeply rooted in the hierarchy of cognitive processing.

Eye gaze measured via eye tracking is a first-order physiological signal, whereas hand annotations (\eg, mouse clicks or bounding boxes) are a second-order manifestation of conscious intent. In other words, eye tracking illustrates the direct link between what the eye fixates on and what the brain reflexively processes. Conversely, hand annotations exist primarily within the top-down task-driven attention space, indicating that visual stimuli are subject to a motor-planning bottleneck. That is, the brain must translate a visual perception into a physical movement, which tends to reflect a low recall for exploration \cite{huang2012user}.

The differences in granularity between the two saliency modalities are also evident from a spatial-temporal perspective. As a continuous stream of data, eye tracking can be susceptible to noise (\eg, saccade movements, jitter, or distraction), whereas hand annotations represent a discrete and comparatively cleaner final result \cite{das2017human}. Beyond these differences, the type of saliency can influence the training of deep learning models. While mouse-contingent paradigms have been proposed as a scalable and cost-friendly substitute for eye tracking \cite{jiang2015salicon}, prior work suggests that hand-based annotations often fail to capture the same spatial distributions as fixations \cite{tavakoli2017saliency}. These findings suggest that models trained on hand annotations may become over-reliant on high-level semantic features while ignoring the subtle, low-level visual cues captured by eye tracking. Although the differences between these two saliency modalities have been explored to some extent, their specific impact on iris PAD remains underexplored.

\section{Experimental Design}
To address our research questions, we perform several experiments testing different saliency methods under the same conditions. In selection of our baseline architecture, we looked to the latest Iris Liveness Detection Competition results \cite{mitcheff2025iris}. While the winning algorithm is a proprietary model, the methods of the runner up team, D-NetPAD and SPAD \cite{sharma2020d, pal2025parametric}, are DenseNet121-based models and thereby motivated our choice of this backbone for our baseline. To further solidify this choice, we trained a closed-set DenseNet121 model on 20,000 full-size iris samples and a test set of 2,000 images for 12 independent runs achieving an average AUROC of 0.999 and an average AP/BPCER 1\% of 0.0068. Motivated by this result, we set out to test robustness to unknown attack classes. We train each configuration in a leave-one-attack-type out manner, where the test set consists of ``live'' samples and the unknown ``attack'' class samples, and perform 12 independent runs per attack type with a total of 7 attack types tested. 
We first conduct a baseline measurement of full-resolution iris PAD classification on the DenseNet121 architecture \cite{huang2017densely, sharma2020d} using standard cross-entropy loss, hereafter simply referred to as ``XENT''.

Next we test a variety of saliency sources in a saliency-guided training paradigm outlined in {Boyd} \etal \cite{boyd2023cyborg} by utilizing the same DenseNet121 architecture with only the saliency source varying across the different configurations. 
Namely, each model is trained with a loss function that consists of cross-entropy and a saliency component.
This saliency component uses the mean squared error of the model saliency with the target saliency in order to bring the model saliency in line with the desired outcome.
Model saliency in our case means the model Class Activation Map (CAM) for a given sample.

Finally, with the emergence of capable foundation models in biometric tasks, we test three classifier configurations using embeddings from DINOv2-Large, DINOv3, and SigLIP2 \cite{oquab2023dinov2, simeoni2025dinov3, tschannen2025siglip}. The details of these remaining configurations are outlined in the following sections.

\subsection{Metrics}
 In accordance with ISO/IEC 30107 we report the Attack Presentation Classification Error Rate (APCER) at Bona Fide Presentation Classification Error Rate (BPCER) threshold of 1\% for each configuration respectively \cite{iso30107}. As we are comparing to the XENT baseline in all cases, specifically for a given model training method we report the difference between its performance and the XENT baseline performance. This does not change that a \textit{lower} value is better for the APCER metric.

 Our final ranking values are averages over all attack types compiled from multiple random variables: the variable for multiple runs per attack type left out, and the average over each category. Due to this we simply report point value ranking and exclude standard deviations. We define a better performance as lower (in the case of APCER).
 
\subsection{Dataset Sources}
The images in our dataset are the same set used in Boyd \etal \cite{Boyd_2022_WACV} and Dowling \etal \cite{11303494} experiments and reflect the following attack types and anomalies along with the bona fide samples and their sources: 
198 real irises \cite{real_iris1,real_artificial_textured_print_24,real_print12,real_diseased,real22,real_textured20,real_textured46,real43,real_textured45}, 
83 irises with textured contact lenses \cite{real_artificial_textured_print_24,real_textured20,real_textured46,real_textured45}, 
57 artificial (plastic eyes and glass prosthesis) eyes \cite{real_artificial_textured_print_24}, 
95 diseased/unhealthy eyes \cite{real_diseased,diseased38}, 
91 printouts  \cite{real_artificial_textured_print_24,real_print12,print21}, 
89 synthetically-generated iris images \cite{synth44}, 
87 printouts of irises with textured contact lenses \cite{real_artificial_textured_print_24}, 
and 65 post-mortem iris images \cite{post40}.

The eye tracking heatmaps were sourced from an eye tracking study conducted by Dowling \etal \cite{11303494} who captured eye gaze heatmaps from a mixture of expert and non-expert examiners.
Acquiring this variety of saliency allows us to perform the first study we are aware of comparing the performance of different human saliency types for saliency-guided training in the task of iris PAD.

\begin{figure*}[ht]
    \centering
    \begin{subfigure}[t]{0.24\textwidth}
        \centering
        \includegraphics[width=\linewidth]{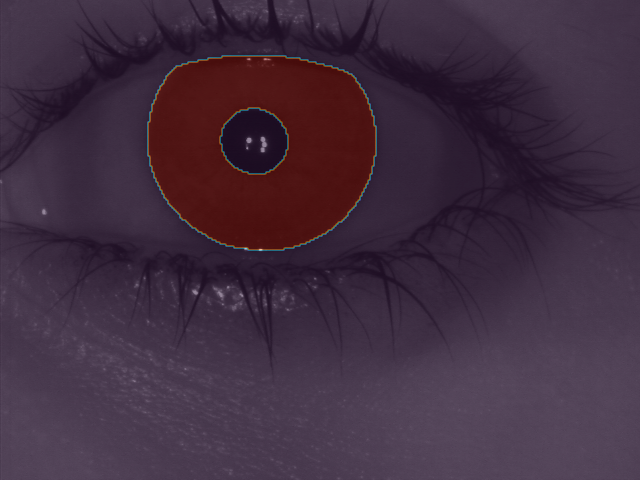}
        \caption{Segmentation Mask of the Iris Region}
    \end{subfigure}
    \hfill
    \begin{subfigure}[t]{0.24\textwidth}
        \centering
        \includegraphics[width=\linewidth]{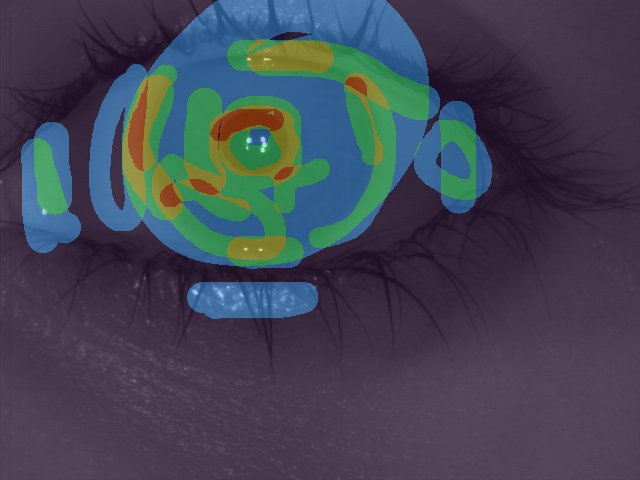}
        \caption{Hand Annotation with Low Entropy}
    \end{subfigure}
    \hfill
    \begin{subfigure}[t]{0.24\textwidth}
        \centering
        \includegraphics[width=\linewidth]{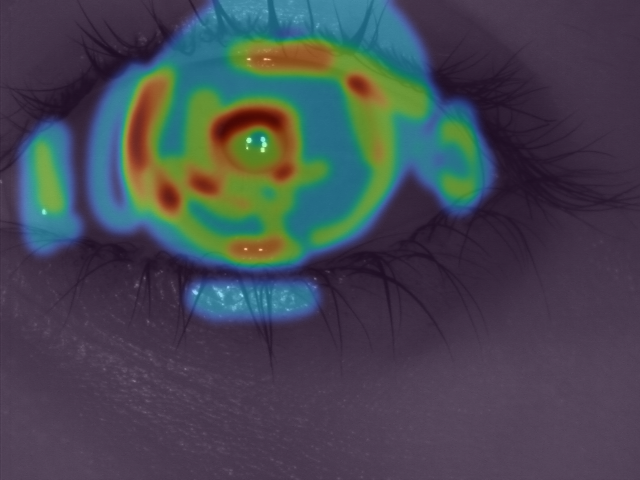}
        \caption{Hand Annotation with Equal Entropy}
    \end{subfigure}
    \hfill
    \begin{subfigure}[t]{0.24\textwidth}
        \centering
        \includegraphics[width=\linewidth]{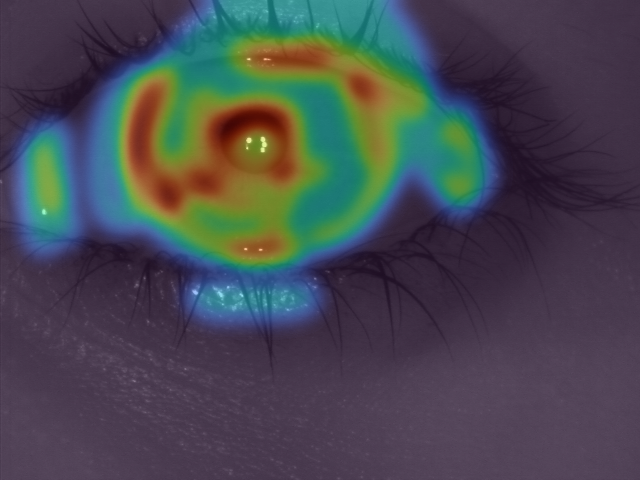}
        \caption{Hand Annotation with High Entropy}
    \end{subfigure}

    \vspace{0.5em}

    \begin{subfigure}[t]{0.24\textwidth}
        \centering
        \includegraphics[width=\linewidth]{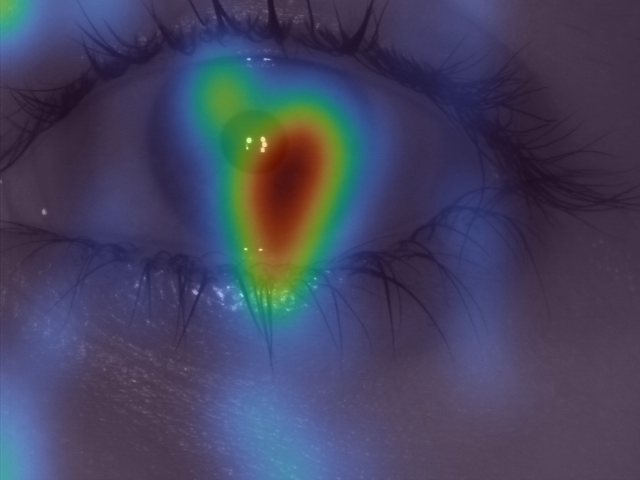}
        \caption{Initial Eye Tracking Heatmap}
    \end{subfigure}
    \hfill
    \begin{subfigure}[t]{0.24\textwidth}
        \centering
        \includegraphics[width=\linewidth]{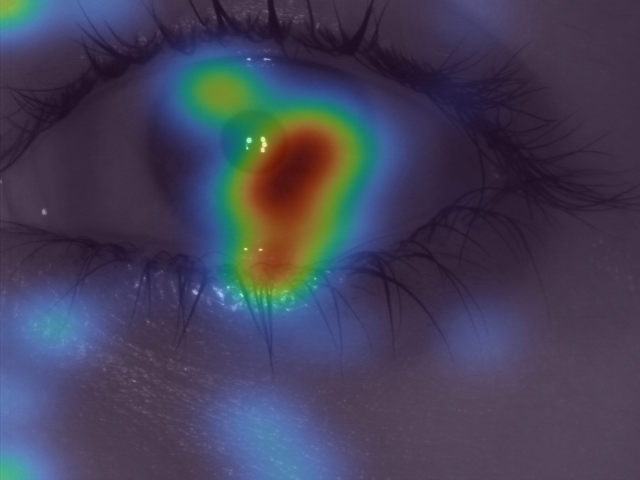}
        \caption{De-noised Initial ET Heatmap}
    \end{subfigure}
    \hfill
    \begin{subfigure}[t]{0.24\textwidth}
        \centering
        \includegraphics[width=\linewidth]{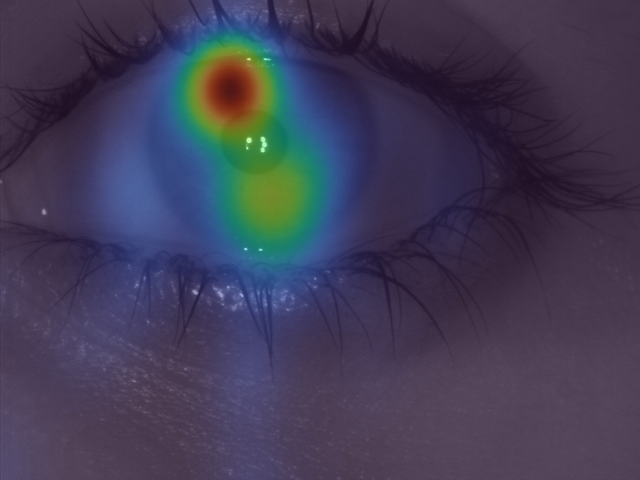}
        \caption{Full Eye Tracking Heatmap}
    \end{subfigure}
    \hfill
    \begin{subfigure}[t]{0.24\textwidth}
        \centering
        \includegraphics[width=\linewidth]{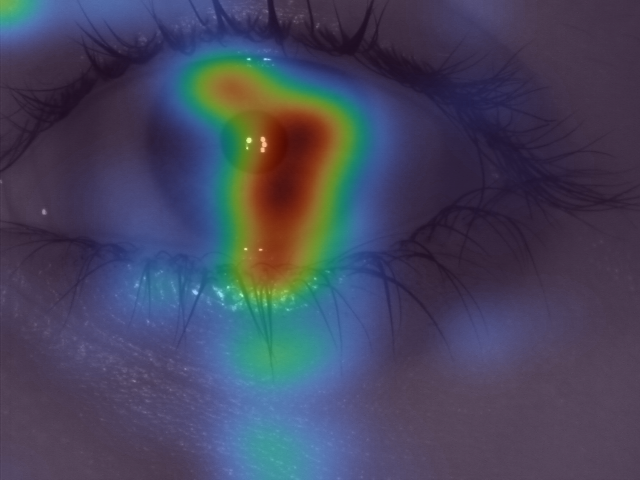}
        \caption{De-noised Full Eye Tracking Heatmap}
    \end{subfigure}

    \caption{Sample salience types captured for the same iris image.}
    \label{fig:four_by_two}
\end{figure*}

\subsection{Saliency Methods}
\subsubsection{Segmentation Masks}
The segmentation masks used in the experiment were generated using a combination of open-source academic iris recognition methods \cite{ND_OpenSourceIrisRecognition_GitHub}.
As stated above, these segmentation masks were used as a part of a loss function to train a model instantiated from DenseNet-121, the full loss function consisting of cross-entropy and an MSE comparison of the model CAM saliency and the segmentation map for a given sample. We refer to this method as the \textit{baseline saliency-based} method as segmentation maps represent the least intricate saliency.

\subsubsection{Hand Annotations}
The human hand annotation saliency we use was published in a human annotation study conducted by Boyd \etal \cite{Boyd_2022_WACV}.
Multiple individuals per sample were tasked with digitally annotating the important features on the sample.
The final heat maps were generated by calculating the per-pixel average number of annotations for a sample, \ie if a pixel was annotated by all individuals who annotated the sample, then the pixel would have a value of 1 and if annotated by no individuals, a value of 0.

We observed that the hand annotations typically have a lower CAM entropy than the eye tracking maps. As CAM entropy can affect model performance \cite{piland2023droid}, we create multiple hand annotation methods at different entropies using Gaussian Blur. Blurring with a kernel of size 10 results in higher entropy on average than the eye tracking data and we refer to training done with these annotations as the \textit{High Entropy} method. Blurring with a kernel of size 5 results in a matching entropy and we refer to training done with these annotations as the \textit{Equal Entropy} method. Using the original annotations with no blur results in a lower entropy than the eye tracking saliency and we refer to training done with these annotations as the \textit{Low Entropy} method.

\subsubsection{Eye Tracking}
Raw eye tracking data collected from both expert and non-expert participants during iris image evaluations was processed into individual saliency heatmaps. For each participant normalized fixation coordinates were first corrected using per-participant polynomial remapping coefficients to account for calibration offset, and a Gaussian Blur was applied to produce a continuous heatmap representation of visual attention. To produce a unified ground truth saliency map for each iris image, individual participant heatmaps were aggregated via a pixel-wise mean across all evaluators, expert and non-expert. 

This pre-processing pipeline was applied for two saliency methods. 
The first we refer to as \textit{Full Eye Tracking}, as it uses captured attention across the full evaluation window.
The second we refer to as \textit{Initial Eye Tracking}, which captured visual attention during a participant's first impression of the iris image. The motivation for examining both phases is to assess whether different visual features are prioritized when forming an initial judgment versus an overall assessment, and to determine which produces saliency maps better suited for saliency-guided training.

\begin{figure}[!t]
    \centering
    \includegraphics[width=\linewidth]{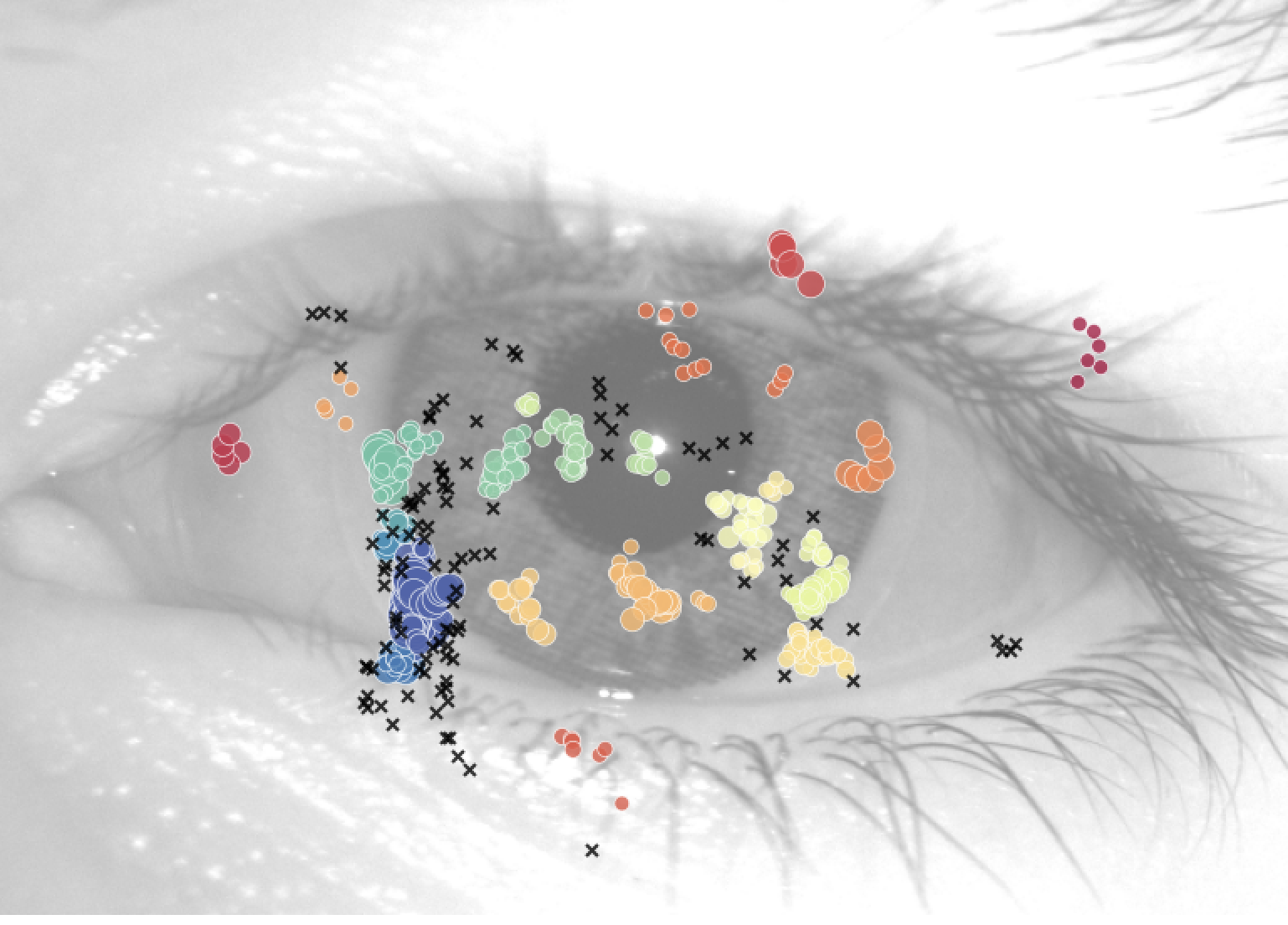}
    \caption{Example of applying HDBSCAN to de-noise the eye tracking data. Clusters made up of valid fixations are distinguished by color and larger marks denote longer fixations, not fixation area. Black crosses indicate fixations marked as noise excluded from the final saliency map.}
    \label{fig:hdbscan}
\end{figure}

However, eye tracking is a continuous stream of gaze points with many commercial models sampling at or above 200 Hz. While high sampling rates of these devices ensure maximum visual features are recorded, these devices are susceptible to noise due to involuntary eye movements such as micro-saccades and eye tremors \cite{martinez2004role, ratliff1950involuntary}. A common technique to reduce noise in eye tracking data, particularly for use in automatic area of interest (AOI) processing, is to apply a clustering algorithm to the fixation data to reduce outlier fixations that do not belong to a cluster according to a minimum cluster size such as DBSCAN or HDBSCAN \cite{ester1996density, campello2013density, eraslan2020best, teng2022visualising}. 

We define two more eye tracking methods using HDBSCAN to de-noise our eye tracking heatmaps. Specifically we apply HDBSCAN with a minimum cluster size of 5 and minimal sample count of 3 to each heatmap (See Fig. \ref{fig:hdbscan}). We refer to training with the de-noised Full Eye Tracking saliency as the \textit{De-noised Full ET} method and training with the de-noised Initial Eye Tracking heatmaps as the \textit{De-noised Initial ET} method. 

\subsection{No Saliency: Foundation Models}
Research has shown that foundation models show a remarkable ability to distinguish both full-size and cropped iris samples in closed-set presentation attack detection settings \cite{sony2025benchmarking, Tapia_FG_2025}. Specifically, Redwan \etal used a variety of foundation models to obtain feature embeddings for iris PAD images and trained three standard classifiers, logistic regression (LogReg) \cite{cox1958regression}, support vector machine (SVM)\cite{cortes1995support}, and support vector machine with radial basis function kernel (SVM-RBF) \cite{scholkopf2000new} on the extracted features. In the task of iris PAD for full size irises, \cite{oquab2023dinov2} suggests that the clear winner was DINOv2. Motivated by these findings, we selected DINOv2-Large and the newer variant DINOv3 \cite{simeoni2025dinov3} to the list of models to test. Finally, motivated by the findings of Piland \etal \cite{piland2026generalist}, whose experiments found that SigLIP2 \cite{tschannen2025siglip} pre-trained embeddings showed good clustering of iris PAD attack types, we select this as our final foundation model to test. We test iris PAD generalization using similar methodology laid out by Redwan \etal by training three standard classifiers, SVM-Linear, SVM-RBF, and a Spatial-CNN, with the embeddings from these three foundation models. Although fine-tuning models can lead to incremental gains in peak performance for specific tasks, we intentionally employ these models as frozen feature extractors. This approach is chosen to ensure consistency with the methodologies established in prior work, avoid high computational overhead, and prevent ``feature drift'' from overwriting the general vision capabilities \cite{tomihari2024understanding, zheng2025learning}.

\begin{table*}[t]
\centering
\caption{Compact $\Delta$ APCER @ BPCER=1\% vs XENT Baseline ($\downarrow$) results are presented.}
\label{tab:compact_apcer_at_bpcer_1}
\small
\renewcommand{\arraystretch}{1.2}
\begin{tabularx}{\textwidth}{@{} l *{7}{>{\centering\arraybackslash}X} | >{\centering\arraybackslash}X @{}}
\toprule
\textbf{Method} & \textbf{Printout} & \textbf{Diseased} & \textbf{Post Mortem} & \textbf{Synthetic} & \textbf{Contacts + Print} & \textbf{Textured Contact} & \textbf{Artificial} & \textbf{Average $\Delta$}\textsuperscript{*}\\
\midrule

\rowcolor{gray!30} \multicolumn{9}{l}{\textbf{Baseline Performance}} \\
DenseNet (XENT) & \textit{0.5540} & \textit{0.9377} & \textit{0.8321} & \textit{0.9831} & \textit{0.2337} & \textit{0.9167} & \textit{0.9868} & \textit{0.7777} \\
\midrule

\rowcolor{gray!30} \multicolumn{9}{l}{\textbf{Segmentation Masks}} \\
Baseline Saliency-based & -0.0522 & +0.0211 & +0.0615 & +0.0094 & +0.0920 & -0.0587 & +0.0088 & +0.0117 \\
\midrule

\rowcolor{gray!30} \multicolumn{9}{l}{\textbf{Hand Annotations}} \\
High Entropy & +0.0256 & +0.0088 & +0.0167 & +0.0169 & +0.0623 & -0.0293 & +0.0000 & +0.0144 \\
Equal Entropy & +0.0879 & -0.0535 & +0.0423 & +0.0094 & \textbf{+0.0172} & -0.0340 & +0.0088 & +0.0112 \\
Low Entropy & -0.0403 & +0.0079 & +0.0282 & +0.0094 & +0.1676 & -0.0082 & +0.0102 & +0.0250 \\
\midrule

\rowcolor{gray!30} \multicolumn{9}{l}{\textbf{Eye Tracking Saliency}} \\
Full Eye Tracking & -0.2766 & -0.0211 & +0.0154 & -0.0665 & +0.0824 & -0.1197 & -0.2354 & -0.0888 \\
Initial Eye Tracking & -0.2701 & -0.0474 & +0.0744 & -0.0431 & +0.1753 & -0.0833 & \textbf{-0.3816} & -0.0823 \\
De-noised Full ET & -0.2894 & -0.0412 & +0.0718 & -0.0927 & +0.2193 & -0.1127 & -0.3553 & -0.0857 \\
De-noised Initial ET & \textbf{-0.3617} & -0.0412 & +0.0551 & -0.1414 & +0.1600 & -0.0599 & -0.3553 & \textbf{-0.1063} \\
\midrule

\rowcolor{gray!30} \multicolumn{9}{l}{\textbf{Foundation Model (No Saliency)}} \\
DINOv2-Large + CNN & +0.3724 & +0.0037 & +0.1010 & -0.0576 & +0.6388 & -0.0073 & -0.0127 & +0.1483 \\
DINOv2-Large + SVM-Linear & +0.4027 & -0.0056 & +0.0892 & -0.0373 & +0.6582 & +0.0043 & +0.0018 & +0.1590 \\
DINOv2-Large + SVM-RBF & +0.4191 & +0.0268 & +0.0940 & -0.0685 & +0.6966 & +0.0479 & -0.0157 & +0.1715 \\
DINOv3 + CNN & +0.3031 & -0.1312 & -0.1780 & -0.1594 & +0.3539 & -0.2010 & -0.0617 & -0.0106 \\
DINOv3 + SVM-Linear & +0.2428 & -0.1047 & -0.1496 & \textbf{-0.2814} & +0.3251 & -0.0549 & -0.0169 & -0.0057 \\
DINOv3 + SVM-RBF & +0.4071 & -0.0649 & -0.0927 & -0.2330 & +0.5127 & -0.0156 & -0.0084 & +0.0721 \\
SigLIP2 + CNN & +0.3216 & \textbf{-0.3820} & -0.1160 & -0.1779 & +0.4142 & -0.0237 & -0.0666 & -0.0043 \\
SigLIP2 + SVM-Linear & +0.2398 & -0.2085 & -0.1428 & -0.1841 & +0.6245 & +0.0042 & -0.0422 & +0.0416 \\
SigLip + SVM-RBF & +0.3171 & -0.1531 & \textbf{-0.2068} & -0.1452 & +0.6574 & \textbf{-0.2049} & -0.0467 & +0.0311 \\
\bottomrule
\end{tabularx}
\vspace{1mm}
\begin{flushleft}
\footnotesize\textsuperscript{*}We use point error estimator in this column (\ie, without calculating and reporting standard deviations) only for ranking the methods, as the average is compiled from random variables of different distributions.
\end{flushleft}
\end{table*}

\begin{table*}[t]
\centering
\caption{Comparison of Saliency-Augmented Training and Foundation Model Performance.}
\label{tab:combined_results}
\vspace{0.05cm}

\begin{minipage}[t]{0.58\textwidth}
\vspace{0pt}
\centering
\caption*{ (a) Pooled APCER at 1\% Threshold (Random-Effects)}
\resizebox{\linewidth}{!}{%
\begin{tabular}{@{}clccc@{}}
\toprule
\textbf{Rank} & \textbf{Method} & \textbf{Pooled} & \textbf{95\% CI} & \textbf{$I^2$ (\%)} \\ \midrule
1 & De-noised Initial ET & 0.6765 & [0.4182, 0.9349] & 97.8 \\
2 & Full Eye Tracking & 0.6918 & [0.4385, 0.9450] & 97.2 \\
3 & De-noised Full ET & 0.6968 & [0.4629, 0.9306] & 98.2 \\
4 & Initial Eye Tracking & 0.7007 & [0.4564, 0.9450] & 98.4 \\
5 & DINOv3 + CNN & 0.7680 & [0.6576, 0.8784] & 98.8 \\
6 & DINOv3 + SVM-Linear & 0.7731 & [0.6334, 0.9127] & 99.3 \\
7 & SigLIP2 + CNN & 0.7756 & [0.6484, 0.9029] & 98.5 \\
8 & Densenet + Baseline & 0.7946 & [0.5416, 1.0477] & 98.1 \\
9 & SigLIP2 + SVM-RBF & 0.8107 & [0.7100, 0.9115] & 98.9 \\
10 & SigLIP2 + SVM-Linear & 0.8214 & [0.7341, 0.9088] & 98.8 \\
11 & Hand Annotations Equal Entropy & 0.8344 & [0.6191, 1.0497] & 97.9 \\
12 & DINOv3 + SVM-RBF & 0.8571 & [0.7611, 0.9531] & 99.3 \\
13 & Segmentation Masks & 0.8677 & [0.6774, 1.0580] & 97.6 \\
14 & Hand Annotations Low Entropy & 0.8769 & [0.7033, 1.0505] & 97.5 \\
15 & Hand Annotations High Entropy & 0.8936 & [0.7316, 1.0556] & 96.9 \\
16 & DINOv2-Large + CNN & 0.9264 & [0.8955, 0.9572] & 99.3 \\
17 & DINOv2-Large + SVM-Linear & 0.9372 & [0.9009, 0.9735] & 99.0 \\
18 & DINOv2-Large + SVM-RBF & 0.9493 & [0.9266, 0.9720] & 99.6 \\ \bottomrule
\end{tabular}%
}
\end{minipage}
\hfill
\begin{minipage}[t]{0.39\textwidth}
\vspace{0pt}
\centering
\caption*{ (b) Robustness Summary (Wins/Losses)}
\resizebox{\linewidth}{!}{%
\begin{tabular}{@{}lccc@{}}
\toprule
\textbf{Method} & \textbf{Sig+} & \textbf{Sig--} & \textbf{Avg $|d|$} \\ \midrule
DINOv3 + CNN & 5 & 2 & 2.469 \\
SigLIP2 + SVM-RBF & 5 & 2 & 3.099 \\
De-noised Initial ET & 4 & \textbf{0} & 1.180 \\
DINOv3 + SVM-Linear & 4 & 2 & 2.105 \\
SigLIP2 + CNN & 4 & 2 & 2.551 \\
SigLIP2 + SVM-Linear & 4 & 2 & 2.255 \\
Initial Eye Tracking & 3 & \textbf{0} & 1.276 \\
De-noised Full ET & 3 & \textbf{0} & 1.188 \\
Full Eye Tracking & 3 & \textbf{0} & 0.854 \\
DINOv3 + SVM-RBF & 1 & 2 & 2.081 \\
DINOv2-Large + SVM-Linear & 1 & 2 & 1.503 \\
DINOv2-Large + CNN & 1 & 3 & 1.801 \\
DINOv2-Large + SVM-RBF & 1 & 3 & 2.191 \\
Hand Annotations Equal Entropy & 0 & \textbf{0} & 0.395 \\
Hand Annotations Low Entropy & 0 & \textbf{0} & 0.318 \\
Segmentation Masks & 0 & \textbf{0} & 0.437 \\
Hand Annotations High Entropy & 0 & 1 & 0.265 \\ 
\multicolumn{4}{c}{\textit{(Methods ranked by significant improvement)}} \\ \bottomrule
\end{tabular}%
}
\end{minipage}
\end{table*}

\section{Results}

\subsection{Statistical Analysis}
For each configuration, 12 independent models per attack-type-left-out were trained for a total of 84 models tested. For each run of 12 models, we report the average APCER at BPCER 1\% threshold. To properly contextualize our results, we first checked Pearson's Correlation Coefficient for all methods against the XENT baseline and confirmed the correlation was near zero for all methods indicating statistical independence. We then employed Welch's unpaired t-test ($p < 0.05$) for all subsequent comparisons to account for unequal variance. Next, we conducted per-attack comparisons of each model configuration against the XENT baseline and using Cohen's \textit{d} effect size, we identify both significant improvement and significant degradation. To evaluate how the results generalize across the entire attack battery, we performed a random-effects meta-analysis using DerSimonian-Laird estimator for between-attack heterogeneity ($\tau^2$). To ensure conservative and reliable estimates given the number of attack categories ($k=7$), we applied the Hartung-Knapp-Sidik-Jonkman (HKSJ) adjustment to the 95\% confidence interval. Methods in table \ref{tab:combined_results} are ranked by their pooled random-effects accounting for within-attack variance and between-attack heterogeneity.

\subsection{Answering RQ1: Does the use of eye tracking saliency in saliency-guided training generalize better to unseen data than other saliency types?}
Table \ref{tab:compact_apcer_at_bpcer_1} shows AP/BPCER 1\% results expressed as each method's improvement above the XENT baseline where each row refers to a method and each column refers to the attack type that was left out during training with a final delta column summarizing the overall performance. Our findings indicate that the De-noised Initial ET configuration shows the largest improvement over the XENT baseline with an improvement of \textbf{0.1063}. The remaining three eye tracking saliency configurations, Full Eye Tracking, De-noised Full ET, and Initial Eye Tracking, improved the XENT baseline by \textbf{0.0888}, \textbf{0.0857}, and \textbf{0.0823} respectively. The remaining saliency-based methods, hand annotations and segmentation masks, failed to surpass the XENT baseline entirely. These results are further reinforced when performing meta-analysis in table \ref{tab:combined_results}. Ranking by weighted mean to account for variance, pooled AP/BPCER 1\% results indicate that eye tracking saliency methods outperformed all other saliency methods with De-noised Initial ET performing the best of all configurations with a pooled average of \textbf{0.6765} with the other eye tracking methods closely behind. \textbf{Thus the answer to RQ1 is affirmative, eye tracking saliency, particularly De-noised Initial Eye Tracking saliency, surpasses other saliency-based methods in detection of unknown attack class for iris PAD.}

\subsection{Answering RQ2: Does saliency-guided training outperform modern PAD solutions utilizing foundational models in detection of unknown attack class for iris PAD?}
When comparing the results of foundation models in table \ref{tab:compact_apcer_at_bpcer_1}, only three of the nine model configurations were able to surpass the XENT baseline, DINOv3 + CNN, DINOv3 + SVM-Linear, and SigLIP2 + CNN. When applying meta-analysis to these results in \ref{tab:combined_results}, the weighted average improves with pooled averages of \textbf{0.7680} for DINOv3 + CNN, \textbf{0.7731} for DINOv3 + SVM-Linear, and \textbf{0.7756} for SigLIP2 + CNN respectively compared to a pooled average of \textbf{0.7946} for XENT. For saliency-based methods, segmentation masks and all hand annotation configurations failed to surpass the pooled average AP/BPCER 1\% of XENT. Eye tracking methods however show strong improvement over XENT with the top four performing configurations for pooled AP/BPCER belonging to an eye tracking method and De-noised Initial ET recording the best pooled AP/BPCER at \textbf{0.6765}. With eye tracking methods improving XENT APCER by 0.0978 or greater and only three foundation model configurations improving the baseline at all, the answer to \textbf{RQ2} is clear, \textbf{saliency-guided training using eye tracking heatmaps outperform foundation models in the detection of unknown attack class for iris PAD.}

\subsection{Observations}
To further contextualize our results, we report on observations from our experiments that may help inform state-of-the-art iris PAD model design. In Table \ref{tab:compact_apcer_at_bpcer_1}, none of the methods were able to improve APCER at BPCER 1\% performance on the ``contacts + print'' attack type as the XENT baseline is already a strong performer at 0.02337. While DINOv2-Large struggled across the board, SigLIP2 and DINOv3 configurations showed strong performance particularly on Post-Mortem, Diseased, and Synthetic samples. When viewing specific configurations in table \ref{tab:combined_results}, DINOv3 + CNN and SigLIP2 + SVM-RBF showed strong performance with both methods improving the baseline for 5 out of 7 attack types and of those improvements, all were significant. Eye tracking methods also made large improvements with all configurations making significant improvements on at least 3 attack types and no significant performance degradation comparative XENT. In particular, De-noised Initial ET significantly increased performance on 4 out of 7 attack types without significantly degrading performance on the remaining three. Furthermore, De-noised Initial ET recorded the best pooled mean AP/BPCER 1\% and the lowest 95\% confidence threshold of all tested methods with similar numbers for all eye tracking configurations. 

While eye tracking saliency performed well across the board, hand annotations did not perform nearly as well. We observe that this aligns with the literature in Sec. \ref{sec:Visual vs Motor Saliency} discussing the motor-planning bottleneck. In Boyd's work \cite{boyd2023comprehensive}, participant's were instructed to highlight 5 regions or features that supported their decision. It is plausible that participants may have highlighted features that were not as important to reach this number, or more than 5 features were needed to support the decision. Additionally, the instruction to focus solely on attack type features may diminish the weight of more subtle textures or a combination of domain features and attack type features that eye tracking more naturally captures.

Finally, we report on the performance between eye tracking methods and foundation models. While foundation model configurations, SigLIP2 + SVM-RBF and DINOv3 + CNN, achieve the highest magnitude of improvement ($|d| scores > 2.469$) with significant improvement on 5 out of 7 attack types, their performance is volatile across attack types as exhibited by their significant performance degradation on the remaining 2 attack types, printouts and printouts of subjects wearing textured contact lens. Conversely, eye tracking methods recorded the highest pooled AP/BPCER 1\% mean values, significantly improved 3-4 attack types, and crucially did not significantly degrade performance on any attack type tested. These findings suggest that eye tracking saliency can effectively regularize the model towards robust features whereas the foundation models tested may overfit to specific high-level features.

\subsection{Limitations and Future Work}
We acknowledge some limitations of this work, which we address in this section. Foundation models were tested in our experiments due to their rising popularity and impressive capabilities in the iris PAD domain. However, we cannot guarantee for any foundational model, including the ones tested here, which iris images, if any, were absorbed into the model training. In terms of performance, the goal of this paper is not to propose an iris PAD method that is better than the state-of-the-art solutions. Instead the main goal is to compare what saliency type helps the saliency-guided iris PAD methods achieve the best generalization to unknown ``attack'' classes. Our results show that collectively eye tracking saliency maps outperformed hand annotations head-to-head while improving the DenseNet121 cross entropy baseline. Since collecting human saliency is expensive and time consuming, and the work presented here is limited by the number of saliency samples, the findings of this work suggest eye tracking saliency is the best candidate for saliency prediction models such as MENTOR \cite{crum2025mentor}, to estimate eye tracking-like (rather than other type of) saliency maps at scale. Future experiments will be conducted to explore eye tracking saliency generation at scale and the impact this has on raising generalization of unknown ``attack'' class to challenge the large-scale application benchmarks. Finally, while our results show that eye tracking is the better candidate to raise generalization for iris PAD, we acknowledge that further exploration in other domains is needed to validate a broader claim. Likewise, our results indicate that de-noising eye tracking maps led to an increase in performance. However, this was largely done to mitigate concerns of a continuous noisy signal commonly associated with eye tracking. Future studies should explore de-noising other eye tracking datasets and consider different clustering parameters depending on the task at hand before a broader claim can be made as to whether de-noising eye tracking saliency maps has a significant impact on saliency-guided training compared to standard eye tracking gaze maps.

\section{Conclusion}
In this study, we compared the performance of methods and forms of saliency, including two distinct human saliency forms, for saliency-guided training in the iris PAD task. Specifically, we made comparisons of APCER at BPCER=1\% against a DenseNet121 cross entropy baseline and reported on the improvement, or absence of, on this baseline. Our findings indicate that the best-performing variant was De-noised Initial ET heatmaps with the best mean pooled AP/BPCER 1\% of \textbf{0.6765} and improving performance on 4 out of the 7 attack types without significantly decreasing performance on the others. The remaining eye tracking methods also showed noticeable improvements with all variants surpassing the mean pooled AP/BPCER 1\% of all other tested configurations including the XENT baseline. Conversely, the majority of the other methods failed to make an impact as neither the hand annotations, nor the segmentation masks, were able to improve over the baseline. Finally, we evaluated three foundation models, DINOv2-Large, DINOv3, and SigLIP2, each with three different classification methods inspired by prior work in closed set evaluations. Our results show that while some configurations of SigLIP2 and DINOv3 made improvements over the baseline, they also actively harmed the baseline in other attack types and ultimately did not exceed the performance of the eye tracking-based methods in mean pooled AP/BPCER 1\%. We conclude that we cannot truly guarantee generalization at the test time for foundation models since their training data is unknown, but their performance increases largely occur on a few attack types and operate in a ``High- Risk, High-Reward'' paradigm. Finally we discuss some limitations and discussed ways for future researchers to adapt this work. To facilitate follow up research efforts, the source codes, model weights, and saliency maps are offered along with this paper.

\section*{Acknowledgments}
This work was supported by the U.S. Department of Defense (Contract No. W52P1J-20-9-3009) and by the National Science Foundation (Grant No. 2237880). Any opinions, findings, and conclusions or recommendations expressed in this material are those of the authors and do not necessarily reflect the views of the National Science Foundation, the U.S. Department of Defense or the U.S. Government. The U.S. Government is authorized to reproduce and distribute reprints for Government purposes, notwithstanding any copyright notation here on.

\bibliographystyle{ieeenat_fullname}
\bibliography{main}

\end{document}